\documentclass[runningheads]{llncs}
\usepackage[T1]{fontenc}
\usepackage{lmodern}
\usepackage{graphicx,verbatim}
\usepackage{hyperref}
\usepackage{svg}
\usepackage{amssymb}
\usepackage{bm}
\usepackage{algorithm}
\usepackage{algorithmicx}
\usepackage{algpseudocode}

\usepackage[most]{tcolorbox}
\usepackage{xcolor}

\renewcommand{\thefootnote}{\fnsymbol{footnote}}

\begin{document}
\title{CORTEX: A Structured Reasoning Benchmark for Trustworthy 3D Chest CT MLLMs}
\titlerunning{CORTEX: A Structured Reasoning Benchmark}

\author{Hashmat Shadab Malik\inst{1}$^{\dagger}$ \and
        Anees Ur Rehman Hashmi\inst{2}$^{\dagger}$ \and
        Numan Saeed\inst{1} \and
        Muzammal Naseer\inst{3} \and
        Salman Khan\inst{1} \and
        Christoph Lippert\inst{2}
        }

\institute{
MBZUAI, Abu Dhabi, UAE \and
Hasso Plattner Institute, Potsdam, Germany \and 
Khalifa University, Abu Dhabi, UAE
}

\authorrunning{Malik and Hashmi et al.}

\maketitle        

\renewcommand{\thefootnote}{\relax}
\footnotetext{$\dagger$ Equal contribution.}

\begin{abstract}
Reasoning in multimodal large language models (MLLMs) has shown strong promise in medical imaging. However, this reasoning is usually free-form text judged only by its final answer, which makes it hard to interpret and verify, especially in 3D radiology, where a diagnosis should be traceable to the evidence in the scan. Existing chest computed tomography (CT) question-answering datasets make this harder by reducing expert radiology reports to answer-only pairs, dropping the reasoning that links findings to conclusions and omitting the patient history clinicians rely on. As a result, reasoning-capable 3D chest CT MLLMs remain out of reach as neither the structured supervision needed to train them nor the protocol needed to verify their reasoning yet exists. We introduce CORTEX (\textbf{C}linically \textbf{O}rganized \textbf{R}easoning and s\textbf{T}ructured \textbf{EX}planation), a structured reasoning benchmark for 3D chest CT. For each question, CORTEX restores the missing reasoning as a four-stage diagnostic trace that mirrors a radiologist's workflow (task understanding, visual observation, diagnostic reasoning, and answer synthesis). We generate these traces using frontier large language models with broad medical and general-domain knowledge, then filter and verify them with a stage-level evaluation protocol that combines automated rubric scoring with expert radiologist review. Crucially, both the reasoning structure and the evaluation rubrics are designed in close collaboration with clinicians. Built on CT-RATE, a large, publicly available chest CT dataset without reasoning annotations, CORTEX comprises 76{,}160 validated reasoning traces across open-ended VQA, closed-ended VQA, and report generation, providing both the structured supervision and the stage-level evaluation protocol needed to build and evaluate trustworthy reasoning models for 3D chest CT. Our dataset and codebase is publicly available at \href{https://github.com/aneesurhashmi/cortex}{Github} and \href{https://huggingface.co/datasets/aneesurhashmi/cortex}{HuggingFace}.

\end{abstract}

\section{Introduction} \label{sec:intro}
\vspace{-0.5em}

Large language models (LLMs) have reshaped natural language processing, achieving strong
performance across diverse tasks through transformer scaling and instruction
tuning~\cite{brown2020language}. Building on this success, MLLMs extend the LLMs by integrating it
with visual encoders, enabling joint vision-language understanding~\cite{liu2023visual,yin2024survey}. The medical community has adapted this paradigm to
clinical imaging, achieving strong performance on medical visual question answering (VQA) and
radiology report generation~\cite{li2024llavamed,wu2025radfm,tu2024towards}. Yet the dominant
paradigm in medical MLLMs is direct response generation, where the model
maps an input to an answer in a single step, with no explicit intermediate reasoning. While this
suffices for simple recognition or retrieval, complex clinical interpretation that requires
integrating heterogeneous visual evidence, reasoning over spatial relationships, and ruling out
competing hypotheses exceeds what single-step generation can reliably support.

A common remedy is chain-of-thought (CoT) reasoning, in which a model produces intermediate
reasoning steps in natural language before committing to a final answer~\cite{wei2022chain}. CoT
can be elicited at inference time through prompting, or taught at training time through
supervision, where models are finetuned on examples that contain explicit reasoning
traces~\cite{hsieh2023distilling}; both have since been extended to multimodal and medical
settings~\cite{chen2025unveiling,ke2025explain,xu2025llava}. In almost all of these, the reasoning
is \emph{free-form}, i.e., unconstrained natural language text with no prescribed structure. In
medicine, this is a critical shortcoming rather than a stylistic one. When a radiologist reads a scan, we accept their diagnosis because they can articulate how specific imaging findings lead to it and defend that reasoning under scrutiny. A model's internal computation, by
contrast, is opaque and its failure modes are difficult to anticipate. A correct answer reached
through an implausible or incomplete path offers no clinical guarantee, since its correctness may
be coincidental rather than systematic. Because free-form CoT imposes no constraints on how
evidence is surveyed, how hypotheses are formed, or how conclusions are justified, its traces vary
widely in structure, depth, and grounding. Even more problematic, holistic evaluation by a
stronger model can prioritize fluency over correctness, rating a confident, articulate chain
highly despite erroneous steps~\cite{tu2026long}.

These problems are most acute in 3D radiology, where interpretation requires coherently integrating evidence across hundreds of interrelated slices that share spatial context
and modality-specific patterns. Existing 3D medical MLLMs demonstrate volumetric understanding through report generation and VQA, but they either answer directly or produce unconstrained reasoning that lacks clinically meaningful
structure~\cite{hamamci2024ct2rep,liu2025argus,li2025towards,lai2026med3d,monon2026lost}, and no radiology specific protocol exists to verify such reasoning at the level of
individual diagnostic steps. Furthermore, almost all existing CT VQA works omit patient history and reason for examination~\cite{zhang2024radgenome,hamamci2024developing,liu2025argus}, even though these are integral to how clinicians reason and determine the correct differential. Without this context, a model must infer from the image alone, its output cannot be checked against the considerations a clinician would actually apply, and
hallucination is encouraged wherever context is critical. Progress toward reasoning 3D CT MLLMs is thus constrained on two fronts; 1) the absence of structured reasoning supervision and 2) the absence of a reasoning specific evaluation protocol. To our
knowledge, no radiology inspired structured reasoning chest CT MLLM yet exists. A key obstacle is that existing CT VQA datasets distill dense radiology reports into \emph{answer-only} QA pairs, discarding the diagnostic reasoning that links findings to
conclusions, the very supervision such a model would need.

Building on this insight, we introduce \textbf{CORTEX} (\textbf{C}linically \textbf{O}rganized
\textbf{R}easoning and s\textbf{T}ructured \textbf{EX}planation), a structured reasoning benchmark
for 3D chest CT that recovers the reasoning discarded by answer-only datasets. CORTEX recasts each sample into a four-stage trace mirroring the radiologist's hypothetico-deductive workflow:
\textit{task understanding}, \textit{visual observation}, \textit{diagnostic reasoning}, and
\textit{answer synthesis}, and reattaches the patient context that prior CT VQA benchmarks omit. We
pair the dataset with a clinician-designed evaluation protocol that scores each reasoning stage
rather than the final answer alone; here we use it to verify trace quality, and it can
subsequently serve to evaluate reasoning models. Both the reasoning structure and the rubrics are
designed in close collaboration with clinicians. We frame CORTEX as a concrete step toward future
reasoning-capable 3D CT MLLMs: it supplies the structured supervision required to train such models
and the stage-level protocol required to evaluate them, with model training and evaluation left to
future work (Sec.~\ref{sec:discussion_and_conclusion}). Our contributions are:

\newtcolorbox{contribbox}{
    enhanced,
    breakable,
    colback=blue!2!white,
    colbacktitle=blue!70!black,
    coltitle=white,
    colframe=blue!60!black,
    fonttitle=\bfseries\large,
    title={\textbf{Key Contributions}},
    arc=3mm,
    boxrule=0.8pt,
    drop shadow={black!20},
    left=3mm,
    right=3mm,
    top=2mm,
    bottom=2mm
}

\begin{contribbox}
\begin{enumerate}
    \item \textbf{Radiologist-inspired reasoning framework.} We introduce
    \textbf{CORTEX}, which structures 3D chest CT interpretation into four clinically
    grounded stages: \textit{task understanding}, \textit{visual observation},
    \textit{diagnostic reasoning}, and \textit{answer synthesis}, mirroring the
    radiologist's hypothetico-deductive workflow and constraining how evidence and
    hypotheses are handled.

    \item \textbf{Clinical context grounding.} We integrate patient history and
    examination indications into both questions and reasoning traces, grounding
    diagnosis in the interaction between findings and presentation while reducing
    hallucination.

    \item \textbf{A clinician-designed evaluation protocol.} Five stage-wise rubrics that score each
    diagnostic step rather than the final answer alone, used here to verify the benchmark and
    reusable to evaluate future reasoning models.

    \item \textbf{A large structured reasoning dataset for 3D chest CT.} We construct a
    dataset pairing volumetric chest CT with multi-stage reasoning traces, comprising
    \textbf{76,177} validated traces (64,224 open-ended, 8,914 closed-ended, and 3,039
    report generation) selected from over 1.9M generated candidates via successive
    rubric-based filtering.
\end{enumerate}
\end{contribbox}

\section{Related Work} \label{sec:related_work}
\vspace{-0.5em}

\noindent\textbf{Reasoning in Medical MLLMs.}
CoT prompting~\cite{wei2022chain} and its use as training supervision~\cite{hsieh2023distilling}
have been adapted to medical MLLMs mainly to expose intermediate rationales. These rationales are
predominantly free-form~\cite{zhang2023multimodalcot}, and where structure exists it is generic:
LLaVA-CoT and chain-of-steps~\cite{xu2025llava,chen2025unveiling} use broad stages, while 2D
methods such as X-Ray-CoT~\cite{wang2025xraycot}, GEMeX-RMCoT~\cite{liu2025gemexrmcot},
ViTAR~\cite{chen2025vitar}, and Citrus-V~\cite{wang2025citrusv} produce region-grounded or
iterative rationales. Volumetric efforts remain largely unstructured: Med3D-R1~\cite{lai2026med3d}
emits free-form rationales for 3D abnormality diagnosis, and 3DReasonKnee~\cite{zhang2025reasonknee}
pairs knee MRI with expert rationales but imposes no staged schema. Across these, the reasoning
trace is shaped by the model rather than the clinician's diagnostic workflow.

\noindent\textbf{Reasoning Benchmarks in Medical Imaging.}
Existing 3D medical imaging benchmarks fall into two groups. \textbf{Outcome-level benchmarks}
score only the final answer: CT-RATE~\cite{hamamci2024developing} and
RadGenome-ChestCT~\cite{zhang2024radgenome} provide large-scale 3D VQA and grounded reports, and
tumor-centric~\cite{chen2025scaling} and spatial~\cite{monon2026lost} VQA probe understanding
through accuracy alone; a parallel modeling line (CT2Rep~\cite{hamamci2024ct2rep},
BTB3D~\cite{hamamci2025btb3d}, COLIPRI~\cite{wald2025colipri}, 3D-CT-GPT++~\cite{chen2024ctgptpp},
DCP-PD~\cite{wang2026dcppd}, Argus~\cite{liu2025argus}) advances tokenization, alignment, or
grounding but is judged holistically, leaving no reasoning to verify. \textbf{Process-level
benchmarks} are closest to ours but target different settings: DrVD-Bench~\cite{zhou2026drvd}
structures clinical reasoning hierarchically but works on 2D slices, and
Med-StepBench~\cite{nguyen2026med} detects step-wise hallucinations in 3D oncological PET/CT via
clinician-verified distractors. CORTEX instead pairs each 3D chest CT volume with structured,
clinician-validated reasoning traces (incorporating patient history where available) and a
five-rubric protocol that scores each diagnostic stage (task understanding, observation fidelity,
hypothesis evaluation, reasoning logic, answer correctness), bringing stage-level verification to
3D chest CT reasoning.
\section{Methodology} \label{sec:methodology}

\begin{figure}[t]
    \centering
    \includegraphics[width=\linewidth]{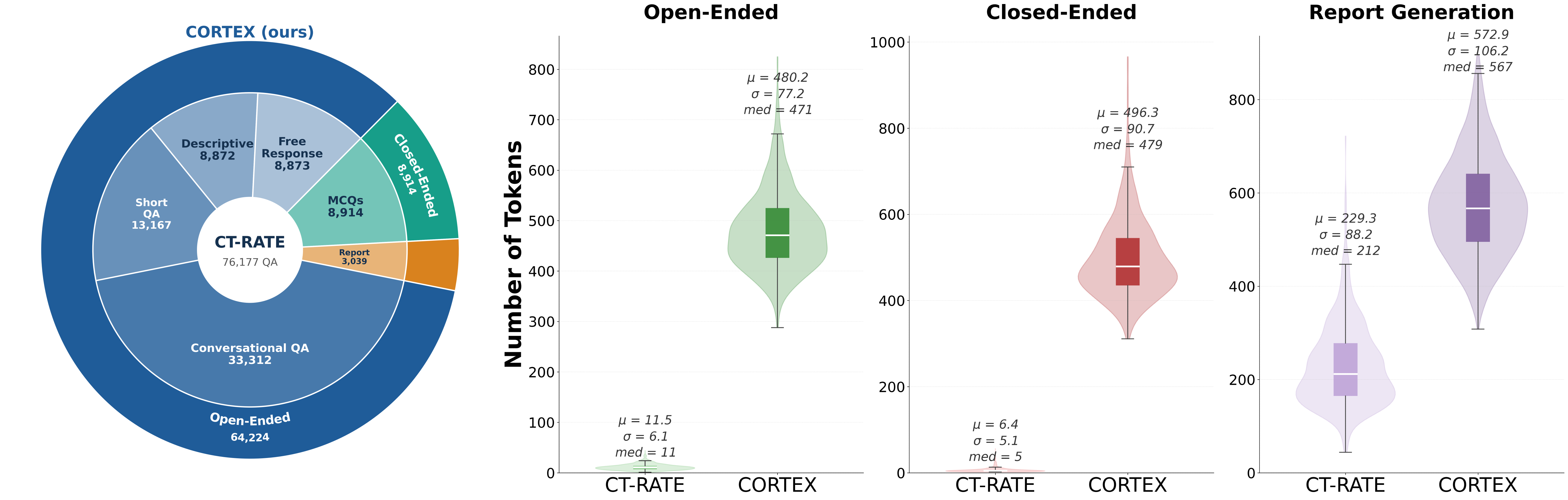}
    \vspace{-1.4em}

    \caption{\textbf{CORTEX restructuring of the CT-RATE validation split.} \emph{Left:} original CT-RATE categories (inner ring) regrouped into three CORTEX VQA types (outer ring), spanning all 76{,}160 QA pairs. \emph{Right:} per-type answer length in tokens for CT-RATE answers vs CORTEX reasoning traces, far longer due to multi-stage reasoning.}

    \label{fig:ct_rate_stats}
    \vspace{-1.0em}
\end{figure}

\begin{figure}[t]
    \centering
    \includegraphics[width=\linewidth]{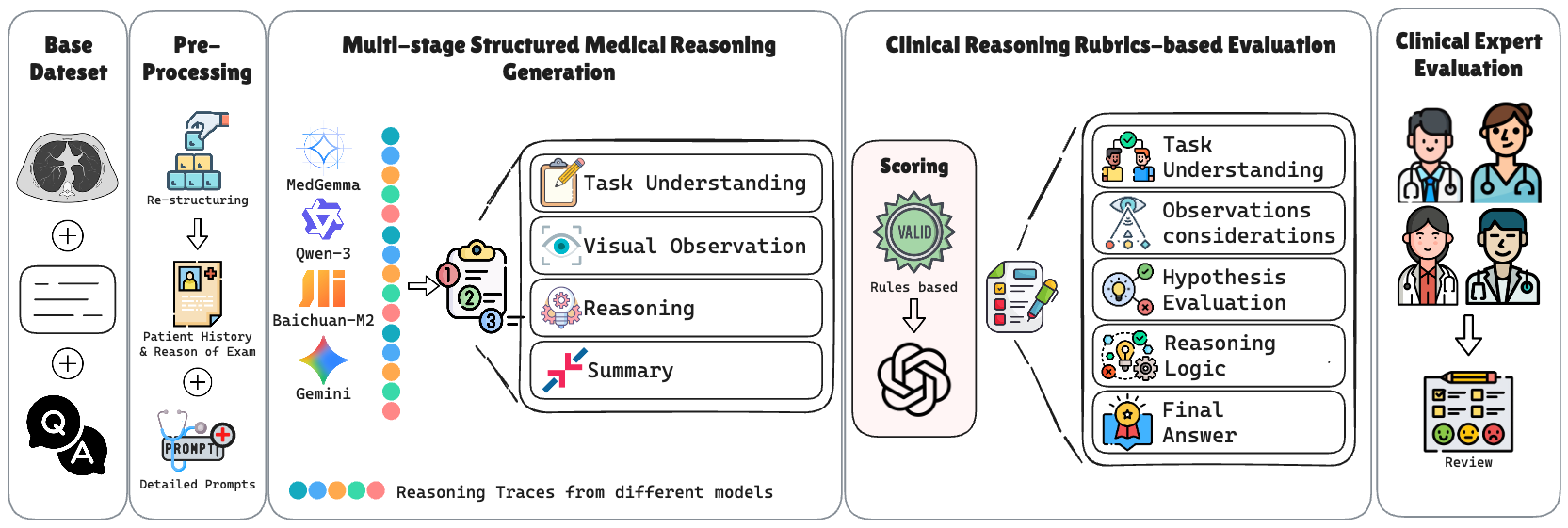}
    \vspace{-1.4em}
    \caption{CORTEX data generation pipeline. Preprocessing restructures the CT-RATE VQA pairs and reattaches clinical context. Using task-specific, clinician-designed prompts, knowledge-rich to frontier LLMs  generate four-stage reasoning traces from the paired reports across multiple temperatures. A rule-based gate and clinician-designed rubrics then score and filter the traces, and the best-scoring trace is retained per sample. Clinicians independently review a random subset of the final dataset.}
    \label{fig:main_fig}
    \vspace{-1.2em}
\end{figure}

CORTEX augments an answer-only chest CT VQA corpus with structured, multi-stage clinical reasoning through a three step pipeline (Fig.~\ref{fig:main_fig}). We (i) restructure and context-enrich the source data, (ii) generate reasoning traces by applying task specific, clinician-designed prompts to frontier LLMs  grounded in the paired radiology report, and (iii) filter the traces with automatic, clinician-designed verification, confirmed by a final expert review. Clinician feedback enters at three points: the generation templates, the verification rubrics, and the independent review of a final subset.  Full clinician-designed prompts, for both reasoning-trace generation and rubric-based validation, along with qualitative samples from different tasks, are available in our codebase.


\subsection{Dataset Preprocessing} \label{sec:dataset_curation}

We build on CT-RATE~\cite{hamamci2024developing}, one of the largest public chest CT corpora, using
its validation split ($1{,}304$ patients, $76{,}177$ QA pairs).
We denote this base corpus as:
\[
\mathcal{D}_{\mathrm{base}} = \big\{\,(V_i,\,\mathcal{P}_i,\,Q_i,\,H_i,\,Y_i)\,\big\}_{i=1}^{N},
\qquad N = 76{,}177,
\]
where, for sample $i$, $V_i$ is the chest CT volume, $\mathcal{P}_i$ its paired report, $Q_i$ the clinical question, $H_i$ the optional clinical context (patient history or reason for examination, $H_i=\varnothing$ if absent), and $Y_i$ the reference answer; volumes and reports are shared across an examination's QA pairs. Preprocessing assigns each sample a reasoning type $\kappa_i$ and a context-enriched query $\bar{Q}_i$, which with $\mathcal{P}_i$ and $Y_i$ form the inputs for reasoning-based generation (Sec.~\ref{sec:reasoning_generation}).

\paragraph{Reasoning-driven restructuring.}
CT-RATE's native categories (conversational, descriptive, free-response, short QA, MCQ, report generation) do not map to distinct reasoning demands. We relabel each sample with a map $\phi$ that sends its category to a reasoning type $\kappa_i \in\mathcal{K}=\{\textsc{oe},\textsc{ce},\textsc{rg}\}$: \textbf{open-ended} ($\textsc{oe}$; conversational, descriptive, free-response, short QA) requires free-text synthesis from question relevant findings, \textbf{closed-ended} ($\textsc{ce}$; MCQ) requires selection from a fixed candidate set, and \textbf{report generation} ($\textsc{rg}$) requires holistic synthesis across volume. Fig.~\ref{fig:ct_rate_stats}\emph{(left)} shows this regrouping for all
$76{,}177$ QA pairs of the dataset.

\paragraph{Clinical context integration.}
Radiology exams are read within a clinical context that informs the diagnosis, yet most CT VQA benchmarks (including CT-RATE) omit it, even though their VQAs derive from the very reports that document it. In $\mathcal{D}_{\mathrm{base}}$, $48.1\%$ of cases ($1{,}462$ examinations) carry such context (e.g., \textit{``myelofibrosis patient, ischemic heart disease''}). We append any available context to the question, $\bar{Q}_i = Q_i \oplus H_i$ (and $\bar{Q}_i = Q_i$ when $H_i=\varnothing$), where $\oplus$ denotes text concatenation.


\subsection{Structured Reasoning Generation} \label{sec:reasoning_generation}

For each sample we synthesize a reasoning trace $R_i$ that derives $Y_i$ from the visual evidence in $V_i$. We treat the report $\mathcal{P}_i$ as a faithful textual surrogate for the volume's visual content and draw the trace from model $g$ at temperature $\tau$,
\begin{equation}
    R_i \sim p_g\big(\,\cdot \mid x_i;\tau\big),
    \qquad x_i = \pi_{\kappa_i}\!\big(\bar{Q}_i, Y_i, \mathcal{P}_i\big),
    \label{eq:gen}
\end{equation}
where $\pi_{\kappa_i}$ is the clinician-designed, task-specific prompt template for type $\kappa_i$. We use a pool $\mathcal{G}$ of frontier open and closed-weight, medical and general-domain LLMs whose large scale pretraining encodes broad medical and general knowledge, sampled over a temperature set $\mathcal{T}\subset[0,1]$ for complementary strengths. Each configuration $(g,\tau)\in\mathcal{G}\times\mathcal{T}$ yields one trace $R_i^{(g,\tau)}$ per sample.

\paragraph{Four-stage reasoning structure.}
We require each trace to follow a four-stage decomposition that makes the diagnostic process explicit and auditable rather than free-form. The design follows the hypothetico-deductive model of diagnosis~\cite{elstein1978medical} and was shaped through structured interviews with our clinician panel, mirroring the radiologist's workflow. Each trace factorizes into ordered stages $\mathcal{S}=(\textsc{tu},\textsc{vo},\textsc{dr},\textsc{as})$, each conditioned on its predecessors:
\begin{equation}
p_g\big(R_i \mid x_i;\tau\big) = \prod_{s\in\mathcal{S}} p_g\big(R_i^{\,s}\mid x_i, R_i^{\,\prec s};\tau\big),
\qquad R_i=\big(R_i^{\textsc{tu}},R_i^{\textsc{vo}},R_i^{\textsc{dr}},R_i^{\textsc{as}}\big).
\label{eq:factor}
\end{equation}
\emph{(\textsc{tu}) Task understanding} restates the question $\bar{Q}_i$, the study type, and any clinical context, anchoring the task before any visual reasoning. \emph{(\textsc{vo}) Visual observation} narrates, in the first person, the volume's findings as recorded in $\mathcal{P}_i$, organized by anatomy. Observations are restricted to question relevant anatomies for
$\textsc{oe}/\textsc{ce}$ and listed exhaustively for $\textsc{rg}$, and anatomy absent from $\mathcal{P}_i$ yields nothing rather than fabrication. \emph{(\textsc{dr}) Diagnostic reasoning} forms candidate hypotheses (the given options for $\textsc{ce}$, or hypotheses induced from $(\bar{Q}_i, R_i^{\textsc{vo}})$ otherwise) and accepts or rejects each using evidence cited from $R_i^{\textsc{vo}}$, yielding an auditable decision path. \emph{(\textsc{as}) Answer synthesis} closes with a concise rationale and a final answer consistent with $Y_i$. The stages are delimited by tags $\langle\textsc{task}\rangle,\langle\textsc{observation}\rangle,\langle\textsc{reason}\rangle,\langle\textsc{answer}\rangle$, making traces training-ready and enabling structural filtering. The resulting traces are substantially longer than the original CT-RATE answers, reflecting the explicit multi-stage reasoning the templates elicit
(Fig.~\ref{fig:ct_rate_stats}\emph{(right)}).


\paragraph{Prompt design.}
We instantiate one task-specific template $\pi_\kappa$ per type, designed from clinician feedback before large scale generation. The clinician panel reviews sample traces and templates (flagging anatomical inconsistencies, implausible hypotheses, or missing instructions), and we revise until the reasoning meets the intended structure at a satisfactory clinical standard. Only finalized templates run over the full corpus, so the reasoning is clinically faithful from the outset.

\subsection{Quality Verification and Trace Selection} \label{sec:validation}

A rule-based gate $\Phi_{\mathrm{rule}}$ first retains only traces whose four-stage tags appear in
the correct order. Surviving traces are scored by an LLM judge $J$ on five clinician-designed,
stage-wise rubrics: task understanding, observation fidelity, hypothesis evaluation, reasoning
logic, and answer correctness. Each is scored on a $1$--$10$ scale with written justifications,
localizing where a trace fails rather than collapsing its quality into a single number.

Empirically, two rubrics (task understanding and answer correctness) consistently saturate at the
top of the scale ($8$--$10$), while the other three vary across traces. We treat these two as a hard
gate $\mathcal{C}_{\mathrm{gate}}$, keeping a trace only if it scores in the $8$--$10$ band on both,
and rank the survivors by the mean of the remaining rubrics $\mathcal{C}_{\mathrm{rank}}$
(observation fidelity, hypothesis evaluation, reasoning logic),
$\bar{q}(R)=\tfrac{1}{3}\sum_{j\in\mathcal{C}_{\mathrm{rank}}} q_j(R)$. Selection is performed
\emph{independently per sample}: for each $i$, we collect every configuration
$(g,\tau)\in\mathcal{G}\times\mathcal{T}$ whose trace passes the rule gate, clears the hard gate, and
meets a quality threshold $\theta$, and keep the single highest-ranked trace among them:
\begin{align}
\mathcal{V}_i &= \Big\{\, R_i^{(g,\tau)} : (g,\tau)\in\mathcal{G}\times\mathcal{T},\;
\Phi_{\mathrm{rule}}(R_i^{(g,\tau)})=1,\; \nonumber\\[2pt]
&\hphantom{{}=\Big\{\,} 
\min_{j\in\mathcal{C}_{\mathrm{gate}}} q_j(R_i^{(g,\tau)}) \ge 8,\;
\bar{q}(R_i^{(g,\tau)}) \ge \theta \,\Big\}, \\[4pt]
R_i^\star &= \operatorname*{arg\,max}_{R \in \mathcal{V}_i} \bar{q}(R),
\qquad
\mathcal{D} = \big\{\, R_i^\star \,\big\}_{i=1}^{N}.
\end{align}
This keeps, for every question, the best-scoring trace across all models and temperatures rather
than committing to a single global configuration.

\subsection{Dataset Generation} \label{sec:dataset_generation}

\begin{figure}[t]
    \centering
    \includegraphics[width=0.96\linewidth]{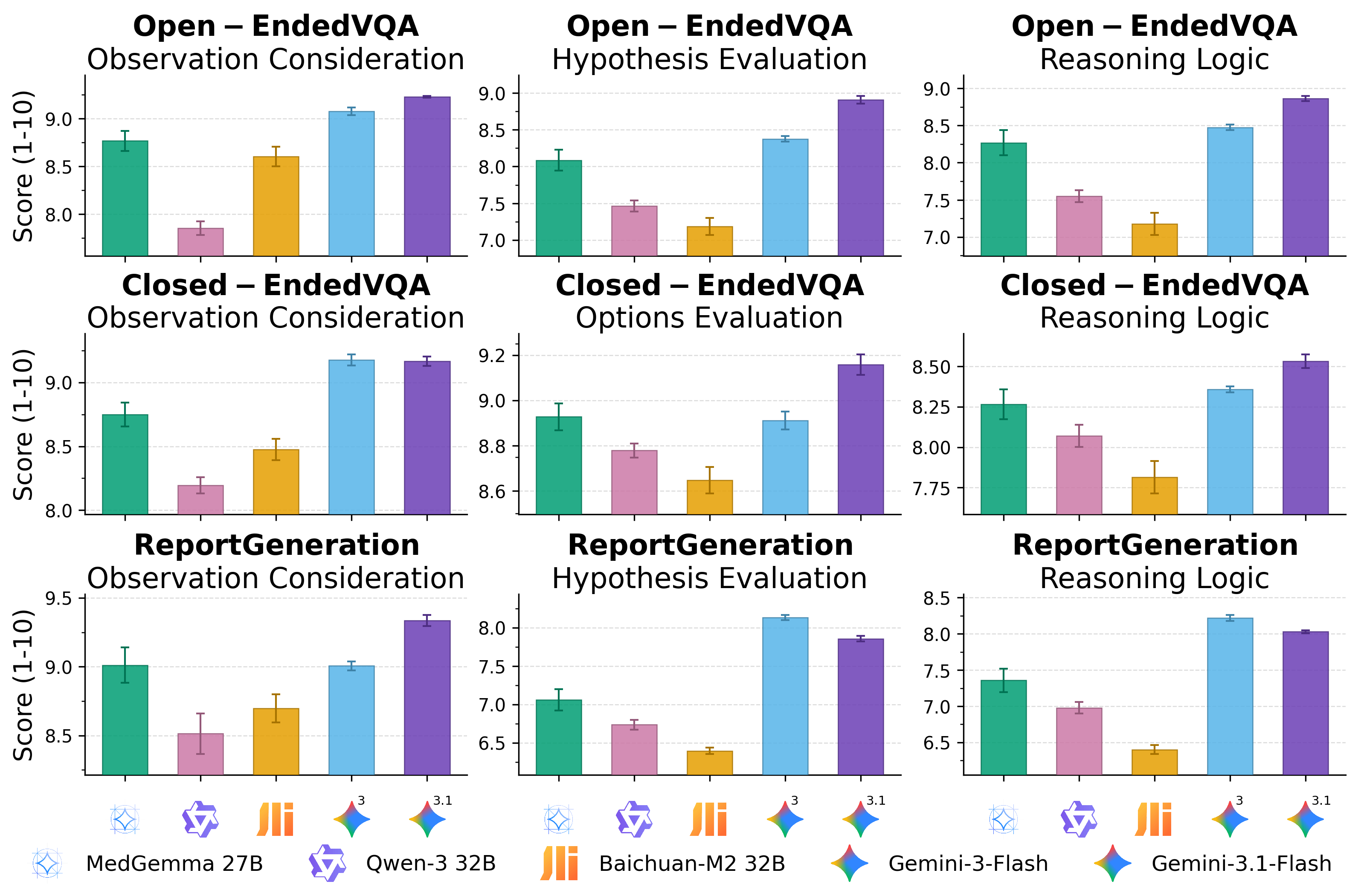}
    \vspace{-0.6em}
    \caption{Rubric-based evaluation of reasoning traces across frontier models. Each bar is a
    model's mean score ($1$--$10$) on the three ranking rubrics (observation fidelity,
    hypothesis evaluation, reasoning logic), averaged over the validation set and the five
    temperatures ($0.0$, $0.2$, $0.5$, $0.7$, $1.0$). The two gate rubrics (task understanding and
    answer correctness) are omitted, as all models score uniformly high. These averages
    characterize overall model behavior and are not used for the per-sample trace selection.}
    \label{fig:model_temp_scores}
    \vspace{-1.2em}
\end{figure}

We instantiate the framework with five frontier LLMs (MedGemma 27B, Qwen-3 32B, Baichuan-M2 32B, Gemini-3-Flash, and Gemini-3.1-Flash), each sampled at five temperatures ($0.0$, $0.2$, $0.5$, $0.7$, $1.0$), producing $1{,}605{,}600$ open-ended, $222{,}850$ closed-ended, and $75{,}975$ report-generation candidate traces. We score these with GPT-5.4-mini as judge $J$ and apply the verification and per-sample selection of Sec.~\ref{sec:validation}, retaining one trace per question. Fig.~\ref{fig:model_temp_scores} reports each generator's mean rubric scores, averaged over the validation set. The resulting dataset $\mathcal{D}$ comprises $64{,}211$ open-ended, $8{,}910$ closed-ended, and $3{,}039$ report-generation reasoning traces,  with each subset divided into training, validation, and test splits in an 80:10:10 ratio. Finally, three board-certified radiologists independently review a random subset of $\mathcal{D}$ ($1{,}000$ questions), labeling each reasoning trace correct or incorrect; inter-rater agreement on correctness reaches $\mathbf{93\%}$, corroborating the automated rubric-based validation.

\section{Discussion and Outlook} \label{sec:discussion_and_conclusion}
\vspace{-0.5em}

We introduced CORTEX, a structured reasoning benchmark for 3D chest CT that pairs each volume with a structured four-stage reasoning trace, reattaches the clinical context prior CT VQA benchmarks discard, and verifies quality through a clinician-designed five-rubric validation and independent expert review. To our knowledge, it is the first resource to make chest CT reasoning explicit,
structured, and verifiable at the level of individual diagnostic stages rather than the final answer
alone. The scope of this work is the construction and validation of this benchmark: at present to our knowledge,
reasoning-capable 3D chest CT MLLMs are underexplored, and progress is held back by two missing resources,
structured reasoning supervision and a reasoning-specific evaluation protocol, both of which CORTEX
supplies. 
Although CORTEX relies on radiology reports as clinically faithful surrogates for the underlying CT volumes, future work can strengthen this paradigm by generating and validating reasoning traces directly from the image data through expert image-grounded annotation, enabling richer visual evidence and reducing dependence on report content alone.
Building on it, our next step is to apply the same generation and verification pipeline at
scale to produce a large reasoning-trace training set, and to use it to train the first
reasoning-capable 3D chest CT MLLMs, evaluated stage by stage with the present protocol, paving the
way toward trustworthy reasoning in 3D radiology.

\bibliographystyle{splncs04}
\bibliography{refs}

\end{document}